\documentclass[10pt,twocolumn,letterpaper]{article}

\usepackage{cvpr}
\usepackage{times}
\usepackage{epsfig}
\usepackage{graphicx}
\usepackage{amsmath}
\usepackage{amssymb}
\usepackage{subfigure}

\usepackage[pagebackref=true,breaklinks=true,letterpaper=true,colorlinks,bookmarks=false]{hyperref}

 \cvprfinalcopy 

\ifcvprfinal\fi
\begin{document}

\title{Beyond Trade-off: Accelerate FCN-based Face Detector with Higher Accuracy}

\newcommand*{\affaddr}[1]{#1} 
\newcommand*{\affmark}[1][*]{\textsuperscript{#1}}
\newcommand*{\email}[1]{\texttt{#1}}
\author{%
Guanglu Song\affmark[1]\thanks{They contributed equally to this work}, Yu Liu\affmark[2]\footnotemark[1], Ming Jiang\affmark[1], Yujie Wang\affmark[1], Junjie Yan\affmark[3], Biao Leng\affmark[1]\thanks{Corresponding author}\\
\affaddr{\affmark[1]Beihang University}, \affaddr{\affmark[2]The Chinese University of Hong Kong}, \affaddr{\affmark[3]Sensetime Group Limited}\\
\email{\{guanglusong,jiangming1406,yujiewang,lengbiao\}@buaa.edu.cn}, \\
\email{yuliu@ee.cuhk.edu.hk, yanjunjie@sensetime.com}
}

\maketitle
\thispagestyle{empty}

\begin{abstract}
Fully convolutional neural network (FCN) has been dominating the game of face detection task for a few years with its congenital capability of sliding-window-searching with shared kernels, which boiled down all the redundant calculation, and most recent state-of-the-art methods such as Faster-RCNN, SSD, YOLO and FPN use FCN as their backbone. So here comes one question: Can we find a universal strategy to further accelerate FCN with higher accuracy, so could accelerate all the recent FCN-based methods? To analyze this, we decompose the face searching space into two orthogonal directions, `scale' and `spatial'. Only a few coordinates in the space expanded by the two base vectors indicate foreground. So if FCN could ignore most of the other points, the searching space and false alarm should be significantly boiled down. Based on this philosophy, a novel method named scale estimation and spatial attention proposal ($S^2AP$) is proposed to pay attention to some specific scales and valid locations in image pyramid. Furthermore, we adopt a masked-convolution operation based on the attention result to accelerate FCN calculation. Experiments show that FCN-based method RPN can be accelerated by about $4\times$ with the help of $S^2AP$ and masked-FCN and at the same time it can also achieve the state-of-the-art on FDDB, AFW and MALF face detection benchmarks as well.

\end{abstract}

\section{Introduction}

\begin{figure}[h]
\centering
\includegraphics[width=0.9\linewidth]{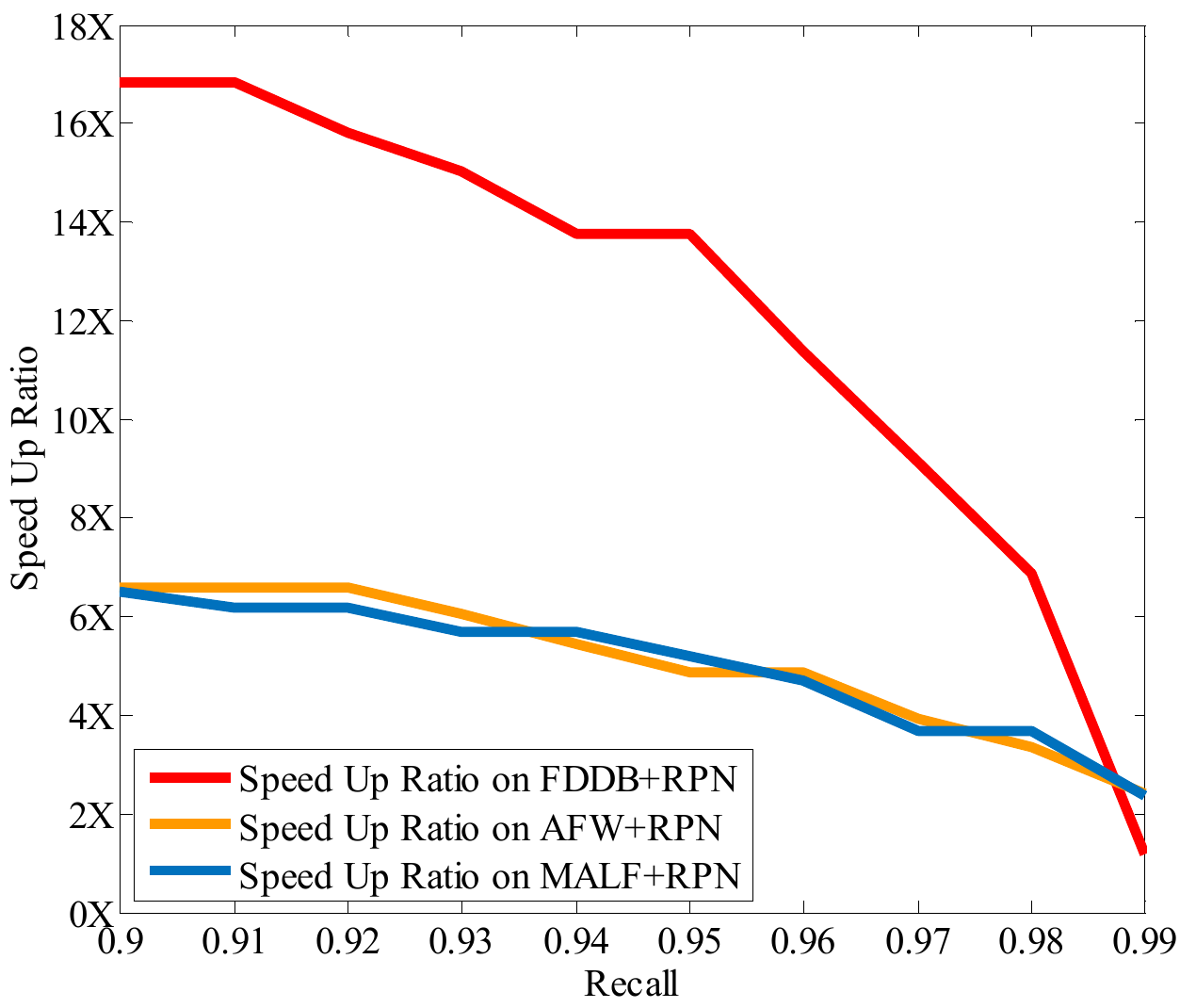}
   \caption{$S^2AP$ is able to considerably diminish the amount of calculation for FCN-based methods such as RPN. It speeds up RPN with image pyramid by 4X on average with 98\% of recall, which indicates the ratio of the number of predicted scales to the number of ground truth and also the number of predicted location proposals to the number of ground truth.}
\label{fig:speed}
\end{figure}

In the field of computer vision, face detection is the fundamental problem for plenty of other applications such as face alignment, recognition and tracking~\cite{sun2014c,sun2015,liu2017rethinking,liu2017quality,yu2016poi}, which is developing faster from the beginning of ~\cite{viola2001rapid} with the emergence of efficient network structure~\cite{he2016deep,szegedy2015going}. But
how to make the face detector both efficient and effective is still a problem.

Face detection is a the special case of generic object detection. Among the top-performing region-based CNN methods, Faster RCNN~\cite{Ren2015Faster} and its variants~\cite{zhu2017cms,jiang2017face,zeng2017crafting,li2017zoom,li2017we} have been developed for face detection task and achieved the state-of-the-art performance. Almost all these methods utilize two-stage mechanism. Proposals are first generated in the first stage, then fed into the second stage with ROI pooling for refinement. However, these methods meet the same problem: the tremendous cost of computation for both extracting features of full image and handling the variance of scales. In order to accelerate the detection pipeline and as much as possible maintain performance, SSD~\cite{Liu2016SSD} and YOLO~\cite{Redmon2016You} adopt single-shot scale-invariant way and try to find a trade-off between speed and precision. ~\cite{Najibi_2017_ICCV,Zhang_2017_ICCV} adopt this manner and detect faces with different scales by using different layers of the network which gains better performance.
Although the scale-invariant method may handle faces in variable scales, it is still unstable to handle a wide range of scale variance, such as from $32\times 32$ to $1024\times 1024$. In view of this situation, the image pyramid is used for handling face scales~\cite{Chen2016Supervised,Li2015A,Hu_2017_CVPR} with a large range of scales and a dense sampling of scales guarantees a higher recall. But the new problems ensue. For one hand, it is hard to choose good layers in image pyramid which include all faces in proper scale. For another hand, the multiple layers in image pyramid with different scales may introduce false alarms, and that will degrade performance. So we will naturally think of the following questions -\emph{What should the sampling scales be?} and \emph{Can we decrease false alarms in image pyramid?}

To better analyze this, we decompose the face searching space into two orthogonal directions, `scale' and `spatial'. Assume that we know the coarse spatial locations and scales of faces, we can pay attention to some specific scale ranges and corresponding locations so that FCN will neglect most of the other space. Then the searching space could be significantly boiled down. Based on this philosophy, scale estimation and spatial attention proposal ($S^2AP$) is proposed to determine the valid layers in image pyramid and valid locations in each scale layer. Furthermore, the masked-convolution operation is used to expedite FCN calculation base on the attention results.

The scheme of $S^2AP$ and masked-convolution operation are comfortable for variable scales, and both convolution operations and scale sampling procedures can be greatly diminished.
$S^2AP$ includes two aspects of attention, `scale' and `spatial'. The former one ensures only the potential layers in image pyramid will be paid attention by FCN and the latter one gets rid of the most background. $S^2AP$ is devised using tiny FCN structure and the computational cost is negligible compared with the later FCN. As shown in Fig~\ref{fig:speed}, FCN-based method such as RPN~\cite{Ren2015Faster} takes advantages of $S^2AP$. When the recall of scale and location is equal to 98\% on FDDB, AFW and MALF, RPN with $S^2AP$ can be accelerated by $4\times$ on average. The `scale' attention further neglects unnecessary scales in the image pyramid which greatly decreases the tremendous time consumption of image pyramid. Further more, experiments demonstrate the FCN-based method RPN with $S^2AP$ greatly diminish false alarms and accomplish the state-of-the-art performance.

To sum up, our contributions in this work are as follows:

1) We propose a novel method named scale estimation and spatial attention proposal ($S^2AP$) that simultaneously estimates the `scale' and `spatial' proposals of the face using the high-level representation in CNN.

2) Masked-convolution operation is implemented for a large reduction of convolution computation in the invalid region with the assist of `spatial' proposals.

3) Our method not only has a significant acceleration effect on FCN-based methods such as RPN but also achieves new state-of-the-art results on FDDB, AFW and MALF face detection benchmarks.

\section{Related Work}

From the CNN-based methods emerging~\cite{Vaillant1994Original} to the breakthrough of approaches~\cite{Viola2004Robust}, the gap between human and face detection algorithms has been significantly reduced. However, large span of face scales and acting convolution operation in the whole image greatly limit the efficiency of face detection.

Many object detection methods have been applied to face detection task such as  Faster-RCNN~\cite{Ren2015Faster} and R-FCN~\cite{dai2016r} etc. The region proposals of the interest area are extracted from RPN and the later stage will further to refine the result of regression and classification. Although these methods can reach the high recall and achieve the satisfactory performance, but the training of the two stages is tedious and time-consuming so that the practical application is hindered. Although ~\cite{Qin_2016_CVPR} designs an alternative joint training architecture for RPN and fast R-CNN, however the single-scale detector requires the image pyramid which also causes expensive computational cost. To break through this bottleneck, YOLO~\cite{Redmon2016You} is proposed to conduct a single stage detection. They perform detection and classification simultaneously by decoding the result from the feature maps and classifying a fixed grid of boxes while regressing them. However, the information of targets with large scale variance is slightly deficient in the high-level feature maps which makes it not easy for multi-scale face detection. SSD~\cite{Liu2016SSD} is proposed for better handling the object with large variation by combining multi-level of predictions from different feature maps. And~\cite{lin2016feature,dollar2014fast} use the feature pyramid to extract the different object information in multi-scale and merge boxes for objects to get high recall. These methods are usually more compatible with multi-scale objects, but the expensive computational cost makes it learn hardly and astatically.

Other researches on face detection are using multi-shot by single-scale detector.
The single-scale detector is configuring for detecting a narrow range scale variance and cannot decode features in other scales. The image pyramid method is proposed for assisting this detector by resizing the image to multi-level scales and then forward the detector. ~\cite{Chen2016Supervised,Li2015A} use the image pyramid to make the single-scale detector capture objects with different scales. When the sampling of scales is dense enough, the higher recall will be achieved. ~\cite{Hu_2017_CVPR} achieves state-of-the-art performance in face detection benchmark based on a proposal network with input pyramid. Although the dense sample of scales will make it possible to detect faces with different scales, but the speed is greatly limited and many different valid samples of scale will bring unreasonable false positives.
\begin{figure*}[t]
\centering
\includegraphics[width=1\linewidth]{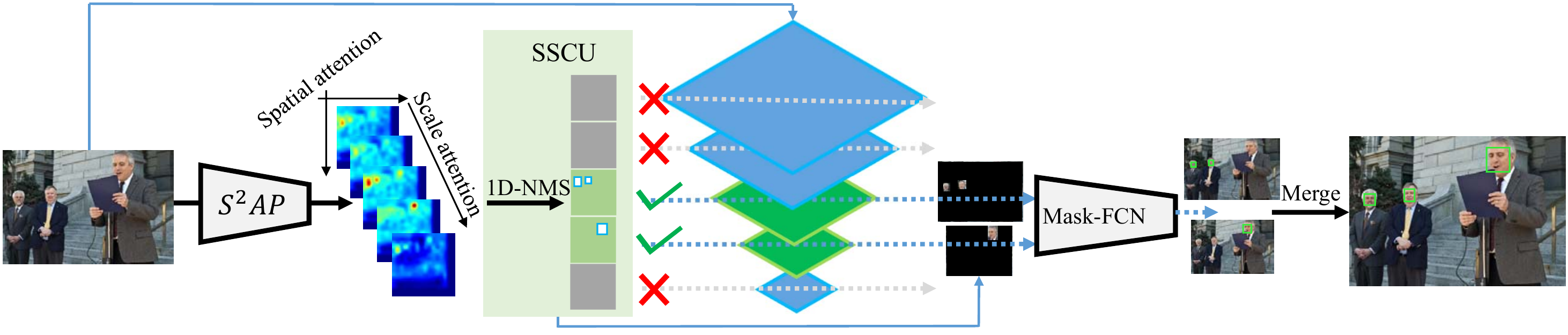}
   \caption{The pipeline of the proposed method. Given an image, it will be fed into the $S^2AP$ with specific scale $448\times 448$ and $S^2AP$ can approximate the potential scales of faces with the corresponding locations. The results of spatial and scale attention are grouped in sixty main feature maps $F=\{F_{1},\cdots,F_{b}\}$($\emph{b} = 1,\cdots,60$). Then, quantitative information precise scale $S$ and meticulous location $C$ are calculated by scale-spatial compute unit(SSCU) and the input $I^{i}$($i = 1,\cdots,length(S)$) and $R^{j}$($j = 1,\cdots,length(C)$) are available for subsequent detection processes. In the last Mask-FCN detector, scale attention helps it zoom in on the image properly and the spatial attention will make the masked-convolution operations and the invalid area will be ignored to better speed up the calculation and effectively dispose of false positives.}
\label{fig:pipeline}
\end{figure*}
Almost all the methods can not escape the bondage to seek a trade-off between the detector's speed and performance. Is there a fundamental method that could accelerate FCN while improving the performance? To analyze this, we decompose the object searching space into two orthogonal directions, `scale' and `spatial'. In order to fully tap the ability of CNN for extracting `scale' and `spatial' information, inspired by~\cite{Hao_2017_CVPR,Chen2016Supervised}, we proposed the scale estimation and spatial attention proposal ($S^2AP$) to better utilize the CNN's ability in approximating the face information of `scale' and `spatial'. Different from SAFD~\cite{Hao_2017_CVPR} using scale information only, location information is further explored for better assisting the prediction of scale while guiding the convolution operator for greatly reducing computation cost and decreasing false positives. ~\cite{yu2017face} also utilizes the scale information for handling variance scales, and the feature map is predicted by the $2\times$ larger than it. The obvious difference from~\cite{yu2017face} is that the scale and spatial information in our framework are highly collaborative work and they will promote each other to make faster and higher accuracy. As the same time, we design the detailed usage of `spatial' for guiding the masked-convolution unlike STN ~\cite{Chen2016Supervised} rough interesting area processing for ROI convolution. The experiments demonstrate $S^2AP$ can greatly accelerate the FCN while deposing the false alarm to further improve performance.

\section{$S^2AP$ with Masked-convolution}

 $S^2AP$ is designed to decrease the cost of computation and false positives. In this section, we depict each component of our system (Fig~\ref{fig:pipeline}). The whole system consists of two sub designs $S^2AP$ and FCN with masked-convolution. $S^2AP$ is a lightweight FCN structure used for fully mining the scale and spatial information of the face. The system will first prognosticate the face scales and location
information included in the image. Then the image will be fed into the latter Mask-FCN with the quantitative information which has been calculated based on the previous scale and spatial information. In the later sub-sections, we will introduce the scale estimation and spatial
attention proposal ($S^2AP$), scale-spatial computation unit (SSCU) and location guided Mask-FCN, respectively. At last, we discuss the adaptability of our algorithm's design over FCN-based method.

\subsection{$S^2AP$}

In order to adequately explore the scale and spatial information of face and take advantage of two orthogonal directions `scale' and `spatial', we devise the delicate lightweight scale estimation and spatial attention proposal ($S^2AP$) which is a fast attention model for pre-detection. The network is a shallow version of ResNet18~\cite{he2016deep} followed by two components, i.e. scale attention and spatial attention.

\textbf{Definition of bounding box.} $S^2AP$ is devised for exploring the information of `scale' and `spatial' so that the misalignment of ground truth bounding box has the obvious effect on training $S^2AP$. Manual labeling of face bounding box is a very subjective task and prone to add noise and in order to retain face size consistent throughout the training dataset, we prefer to derive face box from the more objectively-labeled 5 point facial landmark annotations $(x_i,y_i)(i=1,2,\dots,5)$ which corresponds to the location of \emph{left eye center, right eye center, nose, left mouth corner} and \emph{right mouth corner}. We define $(p_{i},q_{i})(i=1,2,\dots,5)$ for the normalized facial landmark annotations which are formulated as $p_{i}=\frac{x_i-X_{1}}{\emph{w}}$ and $q_{i}=\frac{y_i-Y_{1}}{\emph{h}}$ where $\emph{w}$ and $\emph{h}$ mean the height and width of corresponding manual labeling box and $(X_{1},Y_{1})$ means the top left corner of manual labeling box. The mean point $(mp_{i},mq{i})(i=1,2,\dots,5)$ is computed by averaging all the $(p_{i},q_{i})(i=1,2,\dots,5)$ in dataset. We define the transformation matrix $\emph{T}$ which is a learned similarity transformation between the original landmarks and the standard landmarks as:
\begin{eqnarray}
\left[\begin{array}{c} mp_i \\
mq_i\\
1
\end{array}
\right]^T=
\left[\begin{array}{c} x_i \\
y_i\\
1
\end{array}
\right]^T\emph{T}
\end{eqnarray}
Following this, the consistent bounding boxes can be computed by:
\begin{eqnarray}
\left[\begin{array}{c c} x_{tl} & x_{dr} \\
y_{tl} & y_{dr}\\
1 & 1 \\
\end{array}
\right]^T=
\left[\begin{array}{c c} 0 & 1\\
0 & 1\\
1  & 1\\
\end{array}
\right]^T\emph{$T^{-1}$}  \label{5}
\end{eqnarray}
where $(x_{tl},y_{tl})$ and $(x_{dr},y_{dr})$ mean the top left and bottom right corner of bounding box, respectively.

\textbf{Scale Attention.} The output of $S^2AP$ is a set of feature maps $F$ with $m$ channels (default value of m is 60). Let $F_{b}(\emph{b}\in[1,\cdots,m])$ signifies the feature map which only administrates assigned range of scales. Since the scale of face changes along with the image scaling, we establish a rule to map the face scale to the feature map.
The mapping of face size $x$ and index $\emph{b}$ is defined as:
\begin{eqnarray}
\emph{b} = 10[log_{2}(\frac{x}{\emph{L}_{max}}\times \emph{S}_{max})-4] \label{3}
\end{eqnarray}
where $\emph{L}_{max}$ denotes the maximum value of image's side length and $\emph{S}_{max}$ indicates the predefined longer edge length of the image, which is set to $1024$ in our experiment. The computation of $x$ via the consistent bounding boxes can be formulated as:
\begin{eqnarray}
x=\sqrt{(x_{dr}-x_{tl})*(y_{dr}-y_{tl})}
\end{eqnarray}
When the image is resizing to $\emph{S}_{max}$, faces with scale $2^4$ to $2^{10}$ are equally mapping to $sixty$ main bins $[1,60]$.

\textbf{Spatial Attention.} According to the scale attention, $F_{b}$ express a specific face scale. In the `spatial' attention component, we further explore the information that each coordinate point in $F_{b}$ should contain. Rationalizing a strategy with the assist of the consistent bounding boxes, the value of each coordinate in the $F_{b}$ is formulated as:
\begin{eqnarray}
F_{b}(\frac{(x_{dr}+x_{tl})}{2N_s},\frac{(y_{dr}+y_{tl})}{2N_s})=1, b\in B \label{1}
\end{eqnarray}
where $N_s$ means the stride of the $S^2AP$ network and the face scale defined by $(x_{tl},y_{tl})$ and $(x_{dr},y_{dr})$ corresponds to specific index $b$. We defined $(\frac{(x_{dr}+x_{tl})}{2N_s},\frac{(y_{dr}+y_{tl})}{2N_s})$ for \emph{attention center}. For other coordinates in $F$, the value of them are set to 0. However, simply employing the design above has many drawbacks. It's obvious that the computation of $b$ via Eq.(\ref{3}) are very sensitive to noise and a little deviation from bounding box may cause the difference of $b$. Meanwhile, the interval between the two adjacent scales index $b$ and ${b+1}$ is ambiguous, and its performance drops rapidly with the interval deviation.

Considering the reason above, we utilize a more soft approach for forming ground-truth $F_b$ by comprehensively considering the current index $b$ and its neighbors. For each coordinate value calculated by Eq.(\ref{1}), the value of its neighbor bin can be formulated as:
\begin{eqnarray}
F_{b+i}(x,y)=F_{b+i}(x,y)+(S_{I})^{|i|}, i\in[-4,4],i\neq0 \label{2}
\end{eqnarray}
where $S_{I}=\frac{1}{2}$ which plays the role of extending the effect of current index $b$ to the neighbors and there should be $F_{b+i}(x,y)=min(F_{b+i}(x,y),1)$. We can note that values in $j$-$th$ bin will be enhanced if it is the neighborhood of multi attention centers.

By doing this, the $S^2AP$ is more immune to the interval deviation between adjacent scales since Eq.(\ref{2}) makes border restrictions less stringent. If there appears more than one bounding boxes, these actions are performed for each bounding box.

\textbf{Unified global supervision.} $S^2AP$ unifies the `scale' and `spatial' attention to a single lightweight FCN as shown in Fig~\ref{fig:pipeline}. The output $F$ is treated as the pixel-wise classification problem and is directly supervised by sigmoid cross entropy loss:
\begin{eqnarray}
L=&-\frac{1}{N}\sum_{n=1}^N[p_n(x,y)log\hat{p}_n(x,y)\nonumber \\&+(1-p_n(x,y))log(1-\hat{p}_n(x,y))]
\end{eqnarray}
where $N$ denotes the total number of coordinates in $F$, $\hat{p}_n(x,y)$ is the approximated  response to coordinate $(x,y)$ by the network (normalized by sigmoid function) and $p_n(x,y)$ is the computed ground truth.

Note that during each iteration, the gradient will propagate to each coordinate in $F$ and with the global supervision, the $S^2AP$ can automatically generate scale and location proposal according to features which encode rich information
of face, as shown in $S^2AP$ of Fig~\ref{fig:pipeline}. The global gradient backpropagation not only drives the network to concentrate on the high response scale and location but also instructs the network to distinguish invalid regions and scales.

\subsection{Scale-Spatial Computation Unit}

We have access to scale and spatial information via pre-detection with $S^2AP$ and how can the FCN-based detector make use of aforementioned information? We adopt the Region Proposal Network(RPN) as face detector in our pipeline to verify versatility of $S^2AP$ for FCN-based methods, because RPN is the general expression of FCN and other methods can be extended based on RPN. In order to better embed the scale and spatial information, we employ the \emph{Single-Scale RPN} which has only one anchor with size $64\sqrt{2}$ and has a narrow face size from 64 to 128 pixels. The design guarantees that the overlap between face and anchor is greater than 0.5 for the face scale within the detection range. To capture all faces with different scales, it needs to take multiple shots sampled from an image pyramid.

Define vector of scale information as $S_v=\{max(F_1),\cdots,max(F_b)\}(b\in[1,60])$, where $max(F_b)$ indicates the max value in the feature map $F_b$. We utilize the effective strategy to get the robust information from $S_v$, and scale proposals are obtained by smoothing the $S_v$ and carrying out 1D non-maximum suppression (NMS). The threshold of IOU in 1-D NMS can be regarded as the neighborhood range with $[-4,4]$ which means the \{$S_v^{b+i}$$|$$i\in[-4,4],i\neq0$\} will be abandoned while $S_v^{b}$ has higher confidence. Because of the deviation of network learning, there may be not completely accurate between the ground truth scale and prediction of $S^2AP$. For better handle the prediction gap and make ample use of scale information, we zoom the image as:
\begin{eqnarray}
L_{t}=\frac{2^{6.5}}{x}\times L_{max} \label{8}
\end{eqnarray}
where $L_{t}$ indicates the length of the image's long edges which will be scaled to, $x$ is computed by Eq.(\ref{3}) according to the scale proposals $b$ predicted by $S^2AP$ and $S_{max}$ is 1024 similar with Eq.(\ref{3}). Note that it is beneficial to scale the image to the anchor center size $2^{6.5}$. By doing this, we can guarantee that the target face can also be recalled with overlap greater than 0.5 even if there is a certain deviation $[-4,4]$ between the predicted scale index value and the true scale index value.

Spatial information can be decoded from $F$ according to the scale proposals generated by $S_v$. Taking into account the same situation existing deviation as mentioned above, the final location $C_b$ corresponding to scale index $b$ can be formulated as:
\begin{eqnarray}
C_{b}(x,y)=max(\{F_{b+i}(x,y)|i\in[-4,4]\})
\end{eqnarray}
where $(x,y)$ indicates the coordinates in the feature map. Given the threshold, the regions including faces can be formed from $C_{b}$.

\subsection{Location Guided Mask-FCN}
\begin{figure}[h]
\centering
\includegraphics[width=1\linewidth]{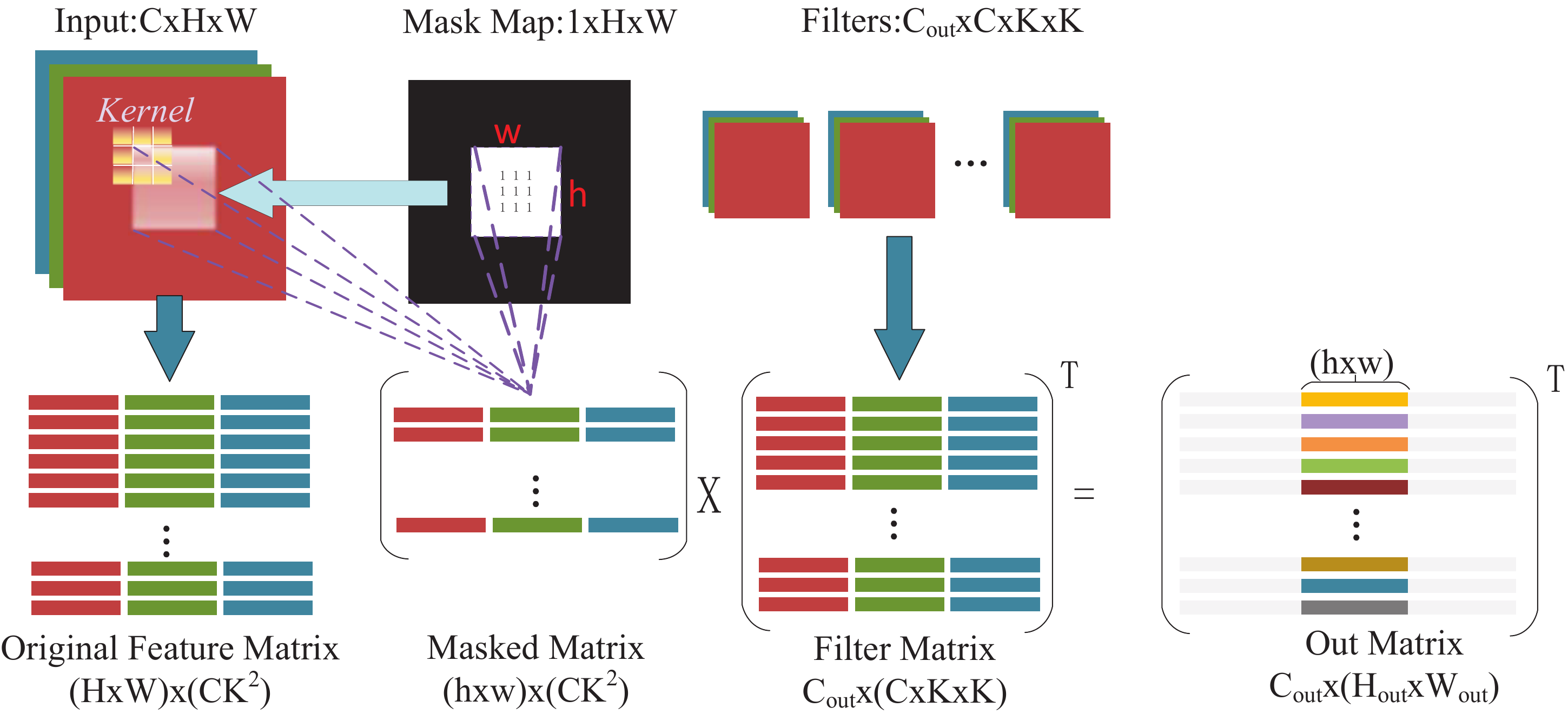}
   \caption{Detail of masked-convolution.}
\label{fig:maskconv}
\end{figure}

The massive computation incurred at test phase of FCN-based methods often limits the practical application. Although the detection stage has been slightly accelerated by FCN, however the cost of convolution computation takes up about more than 90\% of the time in running time, which greatly restricts the speed.

In view of this situation, we implement a more practical approach with the assist of `spatial' information to considerably expedite the speed of FCN-based method. According to the predicted scale $S_v^b$ and its face center location $C_b$, we generate the face regions $((x_{tl},y_{tl}),(x_{dr},y_{dr}))$ and scale it to anchor center size $2^{6.5}$. Besides, in order to retain the context information and alleviate the deviation between predicted location and truth location in $F_b$, we enhance the side length from $l_o=\sqrt{(x_{dr}-x_{tl})*(y_{dr}-y_{tl})}$ to $l_o+2N_s$ where $N_s$ means the stride of FCN. Then, we can generate the location guided mask map where the value is 1 in the potential regions of face and others are 0. Following this, we implement the masked-convolution in the later FCN. The core of this mechanism is that convolution operator only acts on the regions masked as 1, while ignore other regions. As shown in Fig~\ref{fig:maskconv}, we illustrate the input of convolution $I$ with size $C\times H\times W$ and the number of output is $C_{out}$.
On the details of implementation, the input data of original convolution is converted to matrix $D$ with dimensions $(H\times W)\times (CK^2)$ and for the masked-convolution, only the area where the value in the center of sliding window is 1 will get our attention. Then the attention region will be converted to a matrix $D_m=(h\times w)\times(CK^2)$ and $h\times w$ is the number of non-zero entries in the mask map. Similarly, we can use the matrix multiplication to obtain the output $O=D_m\times F$ where matrix $F$ is the filter matrix with dimension $C_{out}\times (C\times K^2)$. Finally, we put each element of $O$ to the corresponding position of the output. Note that the computation complexity of masked-convolution is $(h\times w)\times CK^2\times C_{out}$, therefore we can considerably diminish the computation cost according to the masked-convolution operation guided by the spatial information.

\subsection{Discussion}
\textbf{Excellent lifting power of $\bf{S^2AP}$ to FCN-based methods} The region proposal network is used as our baseline. In our framework, we adopt one anchor with fixed size as the single-scale detector. For handling variable scales of the face, the image pyramid is used via sampling scale densely to make sure each scale face will fall into the detection range of the detector. If only one scale of the face exists in the image, numerous acceleration gains can be obtained with `scale' proposals. Furthermore, another computation that can be greatly accelerated is convolution operation which takes up most of the computing time. `Spatial' proposals can come in handy and masked-convolution can considerably lessen the time of convolution operation through acting on the attention regions while ignoring invalid area. Absence or error in prediction of `scale' and `spatial' proposals will bring performance degradation, therefore, we have added many fine designs aforementioned to solve this problem. Another thing worth noting is that the dense sampling of scales and convolution operations for invalid regions will introduce many false positives. Operating on the specific scale and location will depose the false alarms thereby the performance is capable of further promotion. The latter experiment will prove this strongly.

\section{Experiments}

In the section, we first introduce our setup of experiment and the ablation study to verify the effectiveness of each component in our method. Next, we compare exhaustively with the baseline RPN~\cite{Ren2015Faster} and state-of-the-arts in face detection on popular benchmarks. We also perform experiments on generic object to verify generality and robustness of $S^2AP$.
\begin{table*}
\centering
\begin{center}
\begin{tabular}{l|c c c|c c c}
\hline
   Method & \multicolumn{3}{c|}{RPN} & \multicolumn{3}{c}{RPN+$S^2AP$}\\
\hline
   Dataset & FDDB & AFW & MALF & FDDB & AFW & MALF\\
\hline
   Absolute inference speed (ms)@98\%recall & 95.3 & 95.4 &  94.6 & 14.2& 28.9 & 26.3  \\
   \hline
\end{tabular}
\end{center}
\caption{The proposed algorithm is more computationally efficient than baseline RPN. The Absolute inference speed (ms) at 98\% recall is reported in the table which is performed on NVIDIA P100. The RPN uses the image pyramid.}
\label{table:ratio}
\end{table*}
\subsection{Setup and Implementation Details}
\begin{figure}[h]
\centering
\includegraphics[width=1\linewidth]{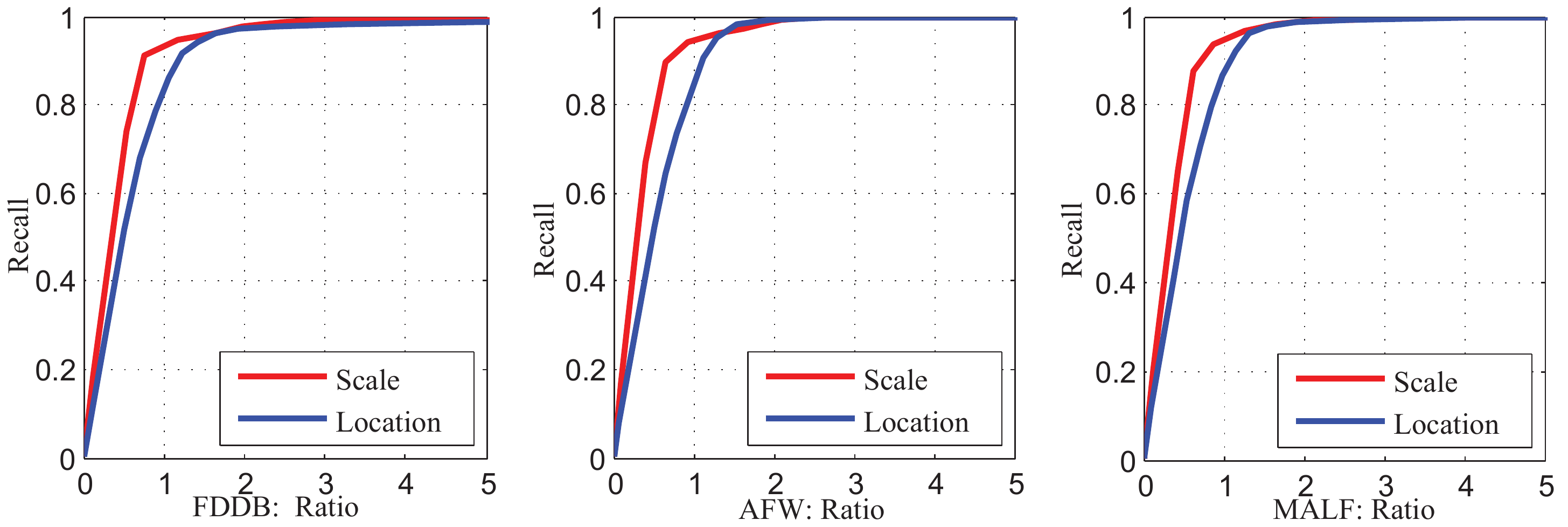}
   \caption{Recall v.s. the ratio of predicted proposals number to ground truth proposals number. This expression can be intuitively responsive to the performance of the network.}
\label{fig:curve}
\end{figure}
\begin{figure}[t]
\centering
\includegraphics[width=1\linewidth]{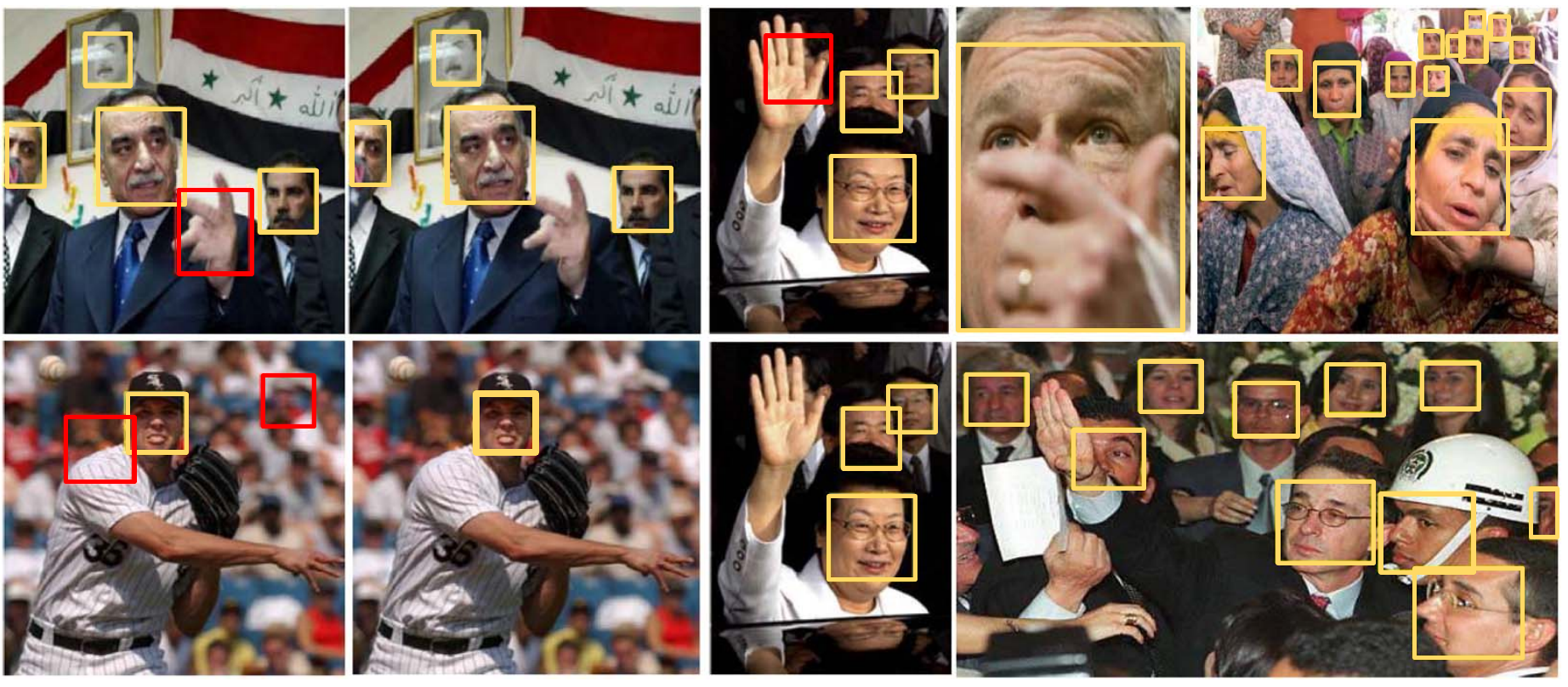}
   \caption{Samples of FDDB detected by RPN and RPN+$S^2AP$. Our algorithm not only deposes many false alarms marked by the red box but also is comfortable with the large scale range of face.}
\label{fig:fddbdet}
\end{figure}
The FDDB~\cite{Jain2010FDDB}, AFW~\cite{Ramanan2012Face} and MALF~\cite{Yang2015Fine} are used for testsets and the configuration is same as~\cite{Hao_2017_CVPR}. Our training set has about 190K images collected from internet and all faces are labeled with bounding boxes and five landmarks. The structure of $S^2AP$ is a lightweight ResNet18 for time efficiency. Similarly, RPN with original ResNet from \emph{input} to \emph{res3b3} as our baseline.
Using shallow and tiny network to be the backbone is faster than using
a whole large network like VGG~\cite{Simonyan2014Very} or ResNet.
In another hand, there is no sufficient receptive field for shallow network to detect large object, so the image pyramid input is significant.
Considering the above aspects, the RPN in our experiments is a single-scale multi-shot detector with fixed anchor size $64\sqrt{2}$. Only the faces in $[64,128]$ can be detected and in training process, we resize the image once to make sure at least one face falls into the scale of $[64,128]$. The training of $S^2AP$ and the RPN detector are initialized by model trained on ImageNet~\cite{Russakovsky2015ImageNet}. In order to ensure the balance of different scale samples while training, we take a random crop on the image to get samples with different scale face. We balance ratio of the positive and the negative to be $1:1$ in training RPN. The base learning rate is set to 0.001 with a decrease of 90\% every 10,000 iterations and the total training iteration is 1,000,000. Stochastic gradient descent is used as the optimizer. In the multi-scale testing stage of baseline, each image is scaled to have long sides of $1414\times 2^k(k=0,-1,-2,-3,-4,-5)$.

\subsection{Performance of $S^2AP$}

The performance of $S^2AP$ is of vital importance to the computational cost and accuracy in the latter FCN-based detector. We validate the performance of the $S^2AP$ on face detection benchmarks and Fig~\ref{fig:curve} demonstrates the overall `scale' and `spatial' recall with predicted scale and location on three benchmarks.
We use \emph{the number ratio} (x, the ratio of total predicted proposals number to total ground truth number) and recall (y, correct predicted proposals over all ground truth proposals) to be the evaluation metric. Compared with~\cite{Hao_2017_CVPR}, our evaluation metric is more precise and the performance is very impressive. We can better recall the most of the ground truth while mistakes are rare at $x=1$. Note that CNN can better explore the scale and spatial information in the high-level representation and this also proves that the network can learn both of the scale and spatial information of the face at the same time.


\begin{figure}[t]
  \centering
  \subfigure[Performance on FDDB with different recall of $S^2AP$. Recall@x means the scale and spatial recall of $S^2AP$.]{
    \label{fig:perform} 
    \includegraphics[height=1.25in]{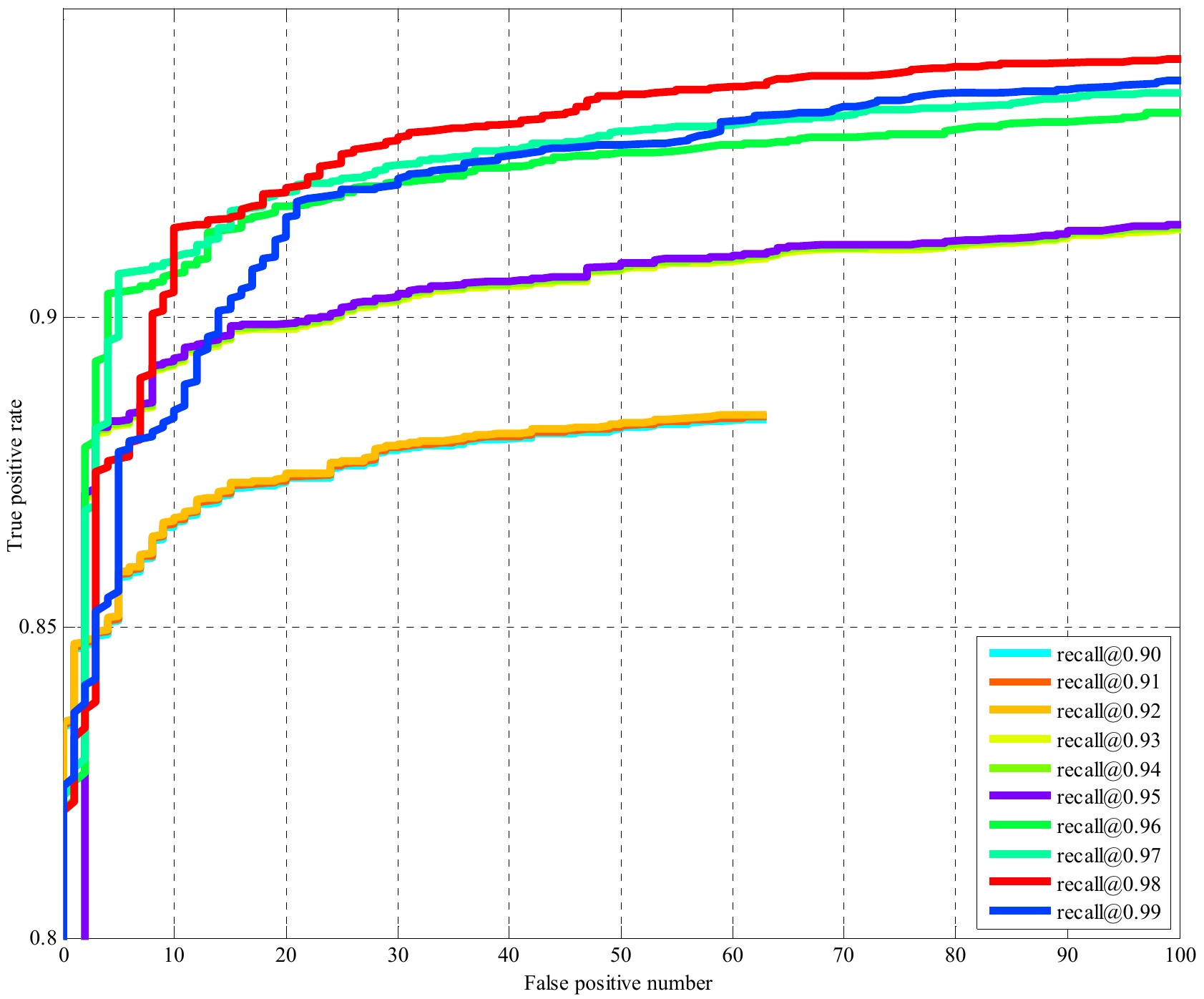}}
  \hspace{0.1in}
  \subfigure[Explore the impact of `scale' or `spatial' attention on detection performance. All the experiments use the same configuration.]{
    \label{fig:abscaleloc} 
    \includegraphics[height=1.25in]{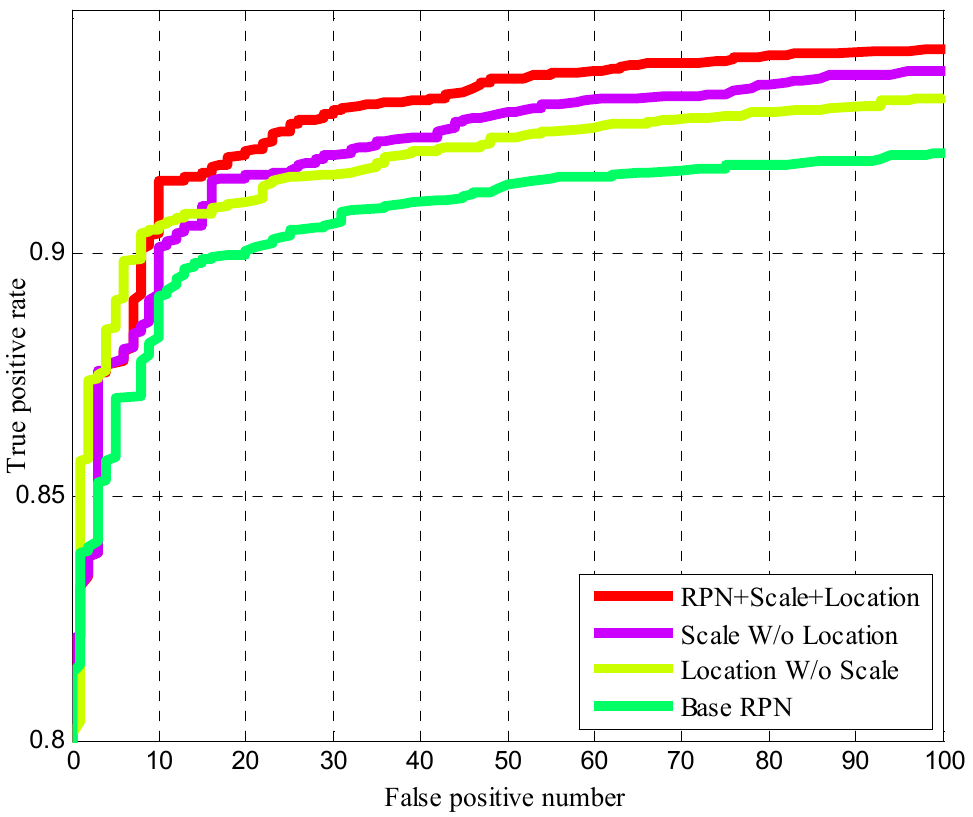}}
      \hspace{0.1in}
  \caption{Ablation study on $S^2AP$.}
  \label{fig:perAndabs} 
\end{figure}

\subsection{Ablation Study on $S^2AP$}

\begin{figure*}[t]
  \centering
  \subfigure[FDDB discrete]{
    \label{fig:subfig:a} 
    \includegraphics[height=1.3in]{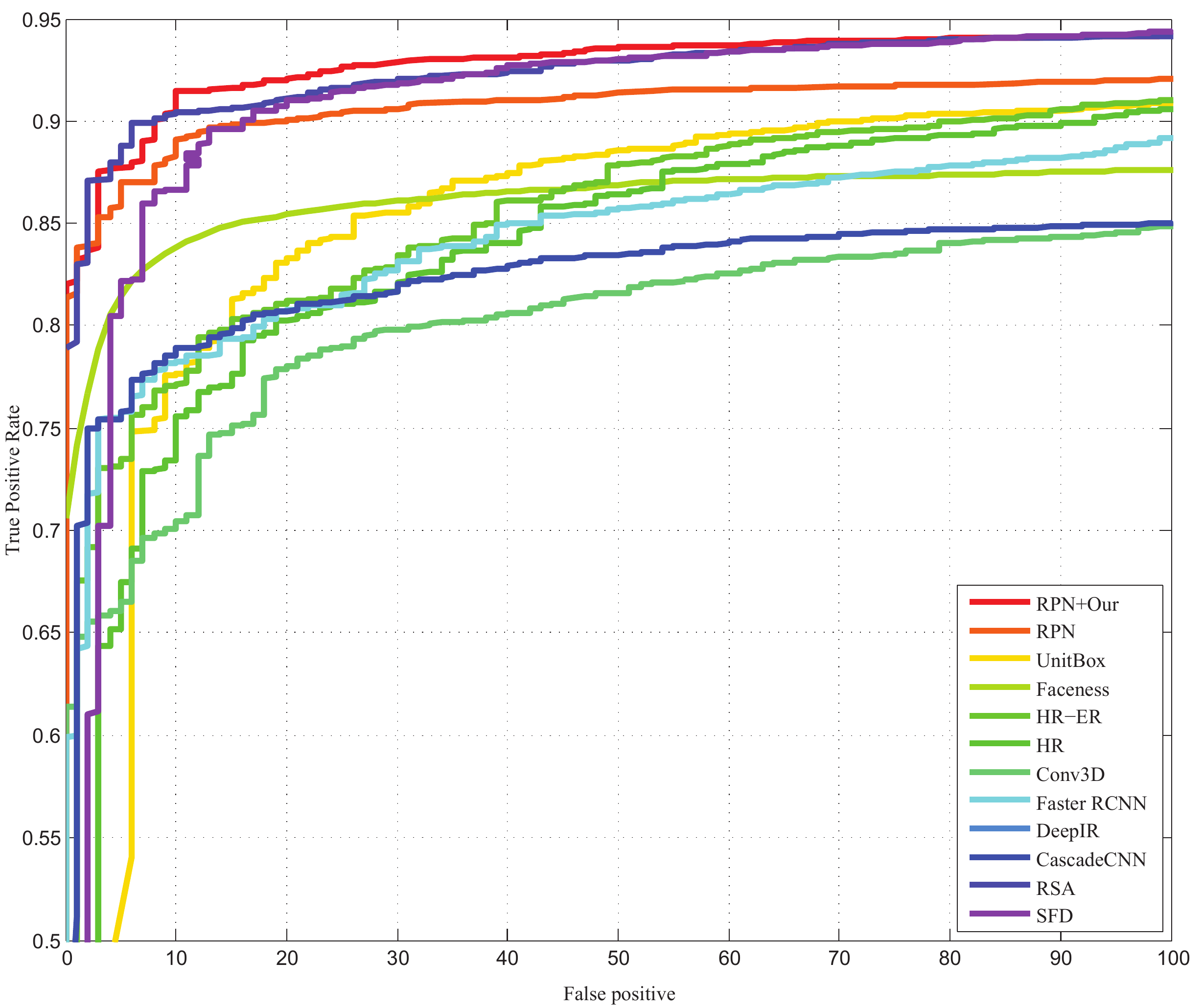}}
  \hspace{0.1in}
  \subfigure[FDDB continuous]{
    \label{fig:subfig:b} 
    \includegraphics[height=1.3in]{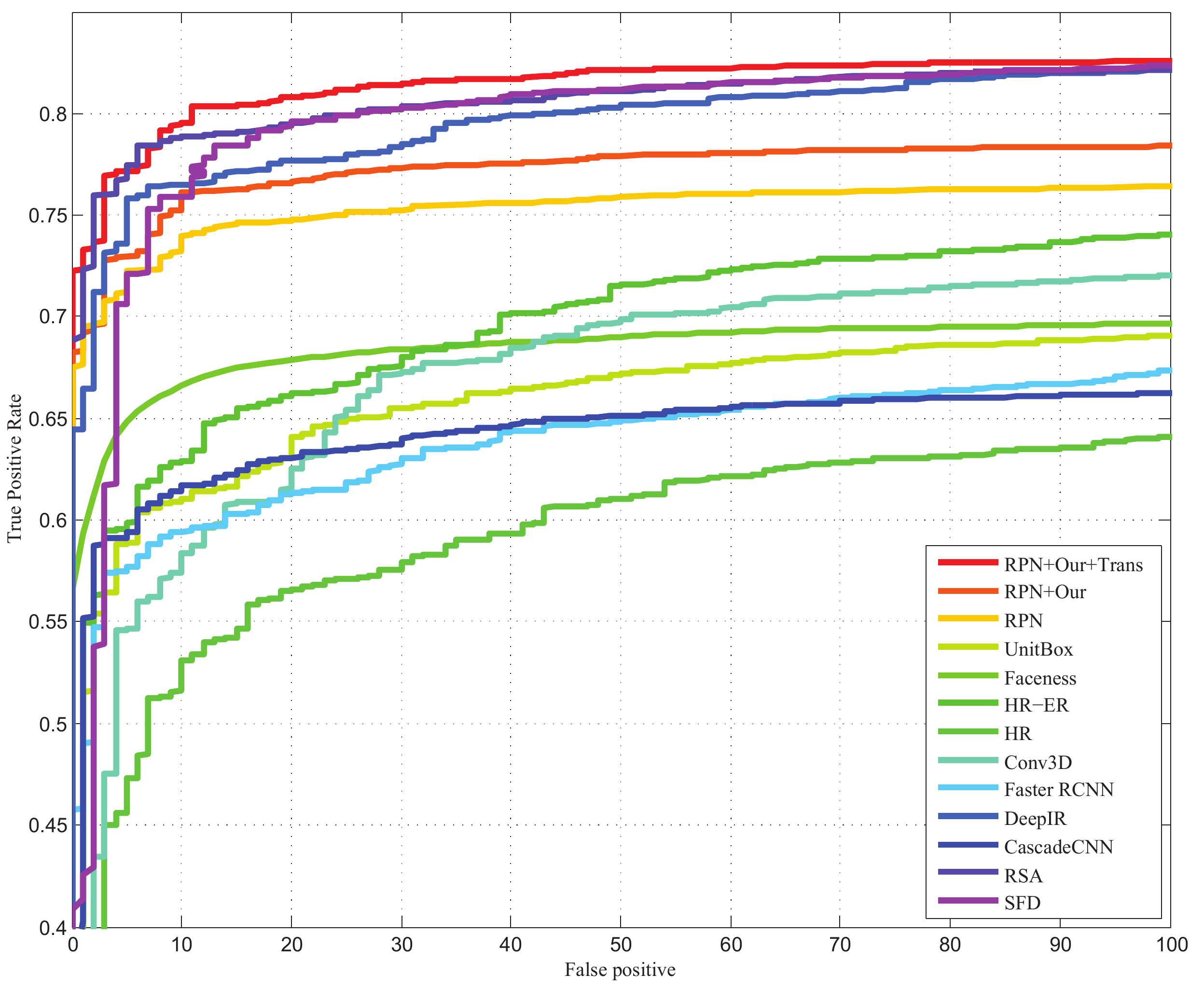}}
      \hspace{0.1in}
       \subfigure[AFW]{
    \label{fig:subfig:c} 
    \includegraphics[height=1.3in]{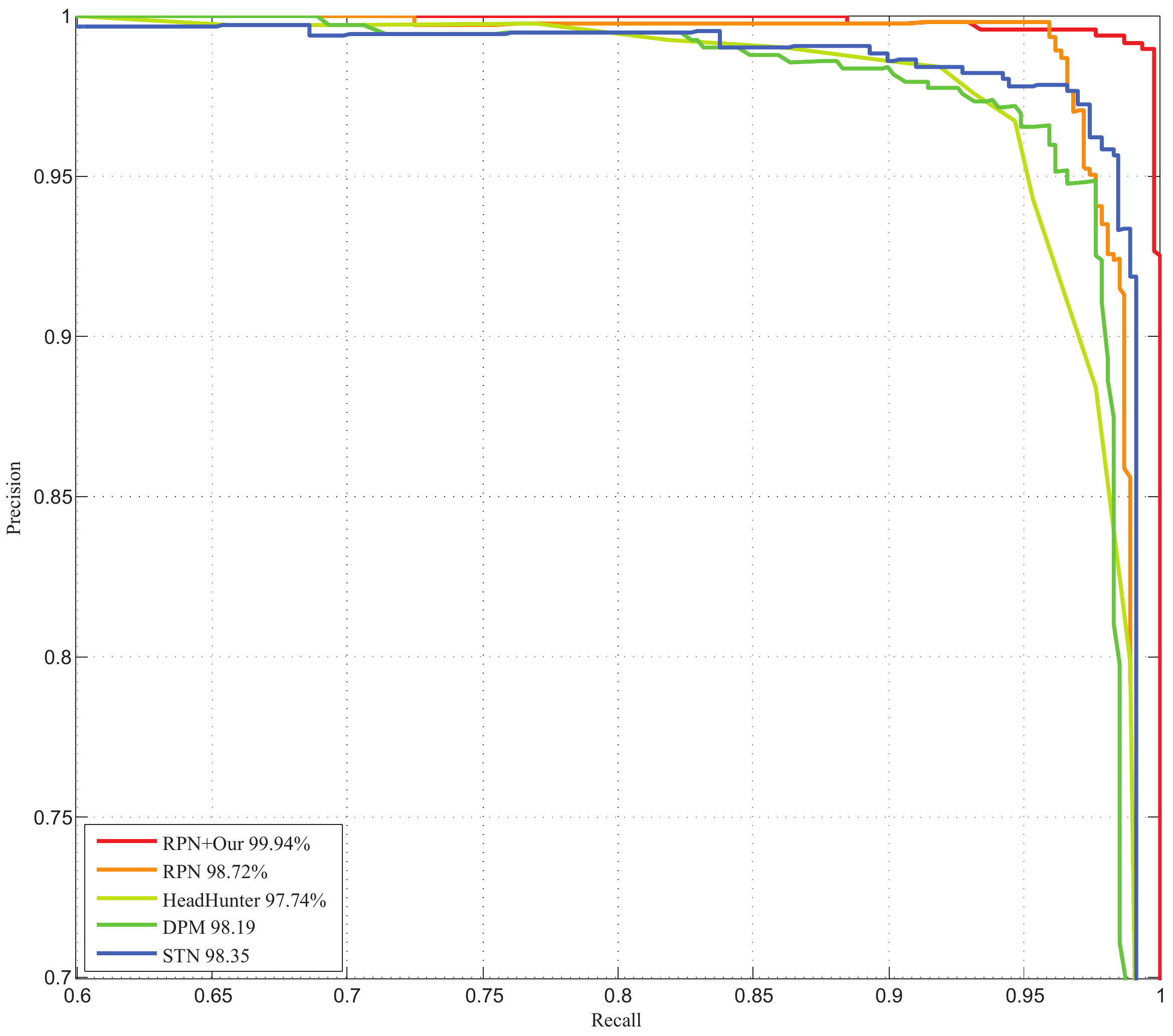}}
    \hspace{0.1in}
  \subfigure[MALF]{
    \label{fig:subfig:d} 
    \includegraphics[height=1.3in]{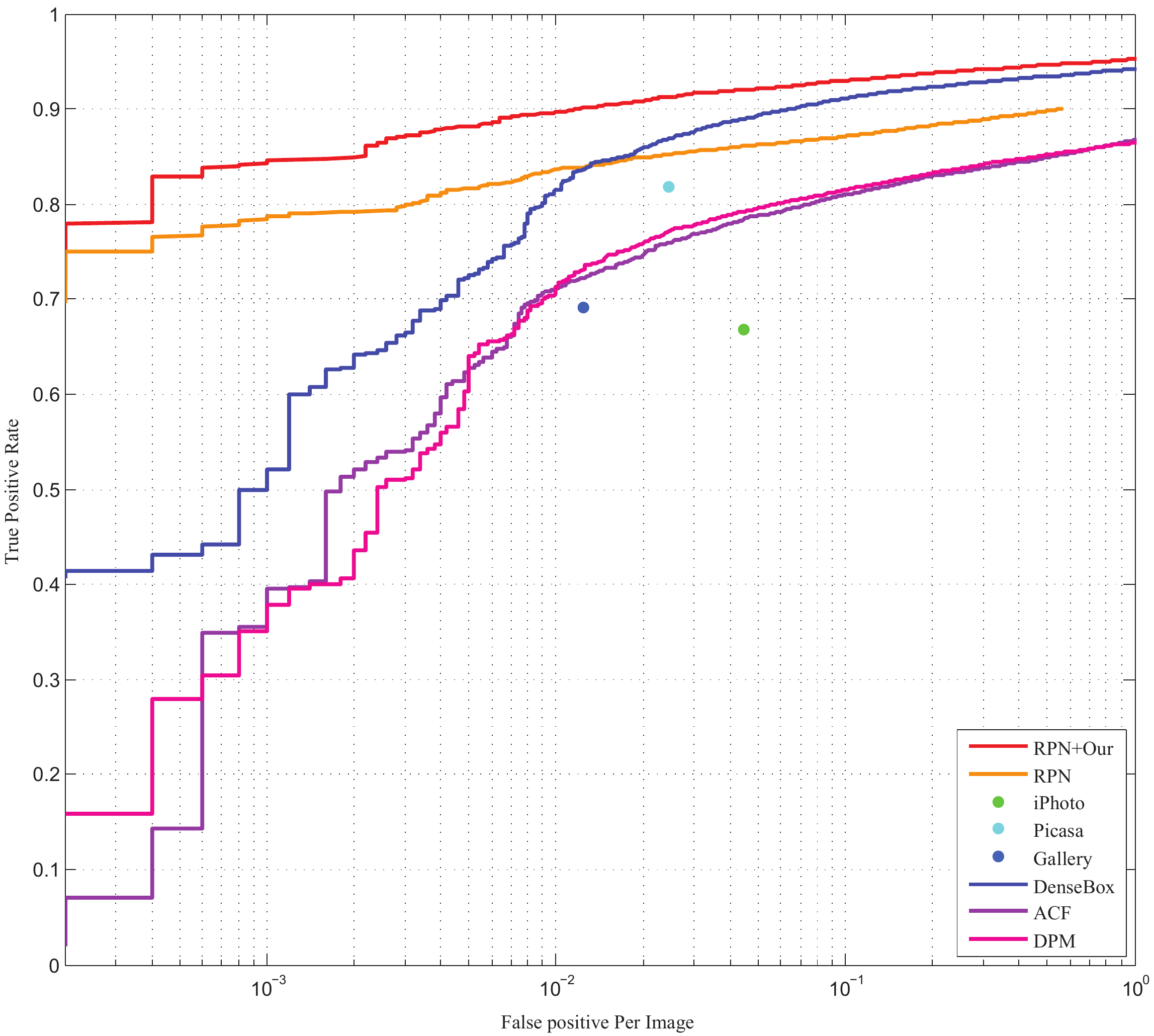}}
  \caption{Comparison to state-of-the-art on face detection benchmarks. The proposed method $S^2AP$ also considerably improves the performance of RPN and outperforms against other methods with an appreciable margin.}
  \label{fig:subfig} 
\end{figure*}


In this section, we perform serial specific designed ablation study on FDDB dataset to detailed prove the effect of $S^2AP$ for FCN-based methods.

First, \emph{acceleration capability}. Theoretically $S^2AP$ can accelerate most of the FCN-based methods with deep CNN architecture whether it is single-scale multi-shot detector or multi-scale single-shot detector. We evaluate the acceleration capability of $S^2AP$ on our baseline single-scale multi-shot RPN with image pyramid and Fig~\ref{fig:speed} shows the different acceleration abilities at the different recall of scale and location proposals. Note that there is a great improvement especially in the lower recall. In the follow-up experiments, we used the threshold of $S^2AP$ at 98\% recall and the acceleration performance at this point is shown in Table~\ref{table:ratio}.

Second, \emph{the ability to improve performance.} The performance of $S^2AP$ has a significant impact on the later stage and we evaluate the ability to improve performance at different recall on FDDB with our baseline. Fig~\ref{fig:perform} demonstrates the performance at different recall of $S^2AP$. Focus on the true positive rate at \emph{false positive number = 50}, the performance at \emph{recall=0.98} is excellent and at \emph{recall=0.92} is showing very low performance because many faces on FDDB are gathered in the same scale, and $S^2AP$ at \emph{recall=0.92} failed to predict this scale which leads to significantly reduced of true positive rate compared with others. Following the better performance at \emph{recall=98\%}, we further compare the performance of RPN and RPN+$S^2AP$, the result is shown in Table~\ref{table:compare}. $S^2AP$ significantly improves the performance of both methods in terms of not only the speed but also the accuracy. Fig~\ref{fig:fddbdet} illustrates that $S^2AP$ can better depose the false alarms and is comfortable with the large scale range of face.

It is particularly important to explore which attention module works, `scale' or `spatial', so we conduct other ablation study on FDDB to explore the ability of each attention. Fig~\ref{fig:abscaleloc} reports the performance on subcomponent. Experiments show that both of the `scale' or `spatial' play their part and promote speed and accuracy more effectively with each other.

\begin{figure}[t]
\centering
\includegraphics[width=1\linewidth]{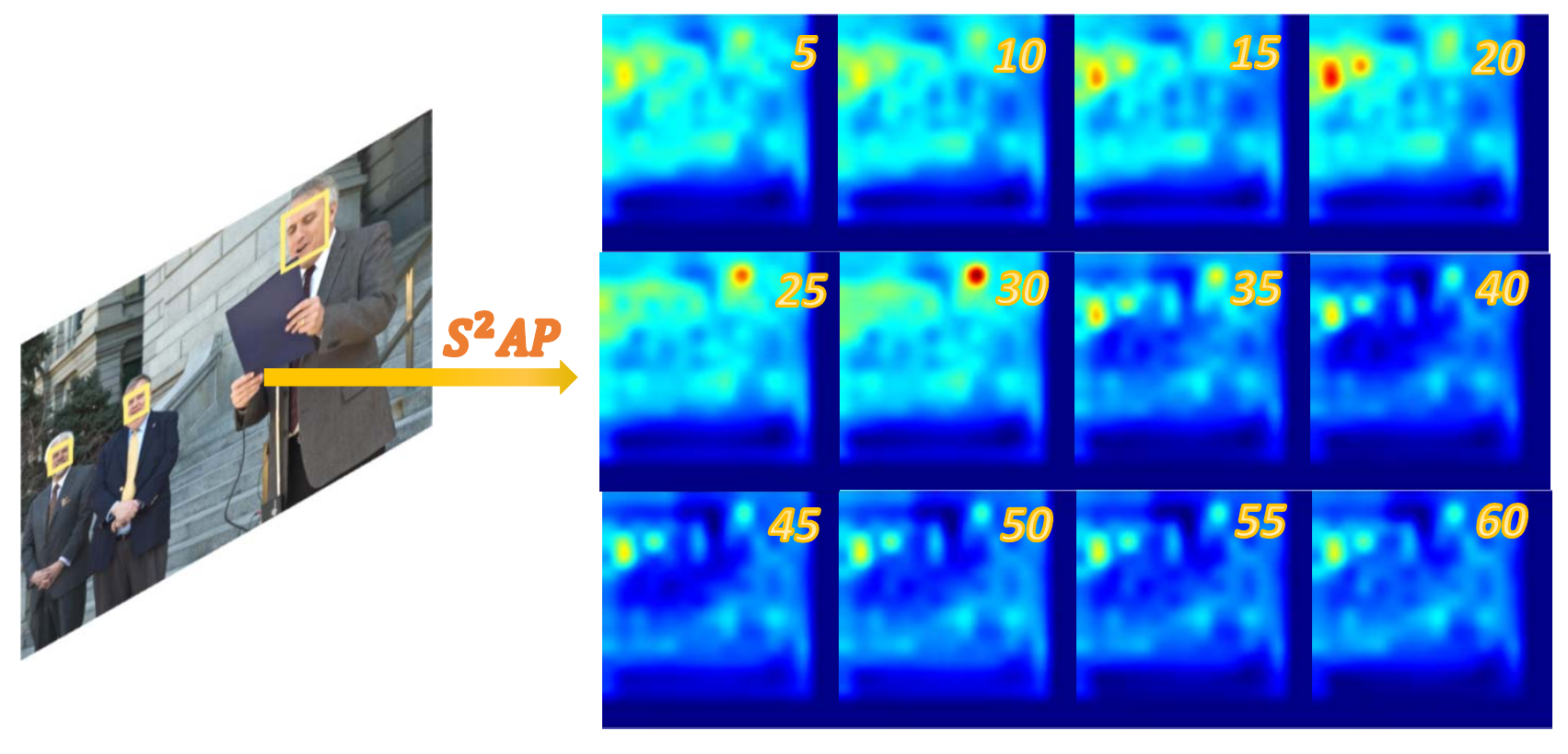}
   \caption{The prediction map generated by $S^2AP$. The number in the upper right corner represents the index $b$ in $F$.}
\label{fig:featmap}
\end{figure}

Figure~\ref{fig:featmap} shows intuitively the prediction map containing `scale' and `spatial' information. The information can be fully excavated from the high response region. It can be  observed the correlation between `scale' and `spatial' that they focus on the target that fall within their control areas in collaboration. By choosing the appropriate threshold, more effective `scale' and `spatial' information can benefit the subsequent detection process.
\begin{table}[t]
\centering
\begin{center}
\begin{tabular}{l|c c c}
\hline
   Method & \multicolumn{3}{c}{FDDB}\\
\hline
   False positive number & 50 &100 &150\\
   \hline
   RPN & 91.39\%&92.03\% & 92.45\%\\
   RPN+$S^2AP$ & \bf{93.59\%}&\bf{94.16\%} &\bf{94.67\%} \\
\hline
\end{tabular}
\end{center}
\caption{The comparison of FCN-based method with $S^2AP$. The threshold of $S^2AP$ is determined by $S^2AP$ recall=98\%.}
\label{table:compare}
\end{table}

\subsection{Comparing with State-of-the-art}

We conduct face detection experiments on three benchmark datasets FDDB, AFW and MALF and we compared with all public methods~\cite{huang2015densebox,mathias2014face,yang2014aggregate,Hu_2017_CVPR,Li2016Face,Yang_2015_ICCV,Ren2015Faster,Chen2016Supervised,Li2015A,sun2017face,yu2017face,Zhang_2017_ICCV,yu2016unitbox} and so on. We regress the annotation with 5 facial points according to Eq.~\ref{5} and Fig~\ref{fig:subfig} demonstrates the comparison. As can be seen from the figure, our method outperforms all previous methods by a appreciable margin. On AFW, our algorithm achieves an AP of 99.94\% using \emph{RPN+$S^2AP$}. On FDDB, \emph{RPN+$S^2AP$} recalls 93.59\% faces with 50 false positive higher than~\cite{yu2017face} which also utilizes the scale information and on MALF our method recalls 77.92\% faces with zeros false positive. Note that the shape and scale definition of bounding box on each benchmark varies. In particular, the label of the FDDB is an ellipse which is different from the standard of the bounding boxes we regress according to landmarks. In order to better adapt the standard of FDDB, we learn a transformer to transform our bounding boxes to the target and \emph{RPN+$S^2AP$+Trans} in the setting of FDDB continuous significantly enhances performance.

\subsection{Generality of $S^2AP$ on Generic Object}

\begin{figure}[t]
  \centering
  \subfigure[Performance on VOC]{
    \label{fig:subfig:a} 
    \includegraphics[width=1.5in]{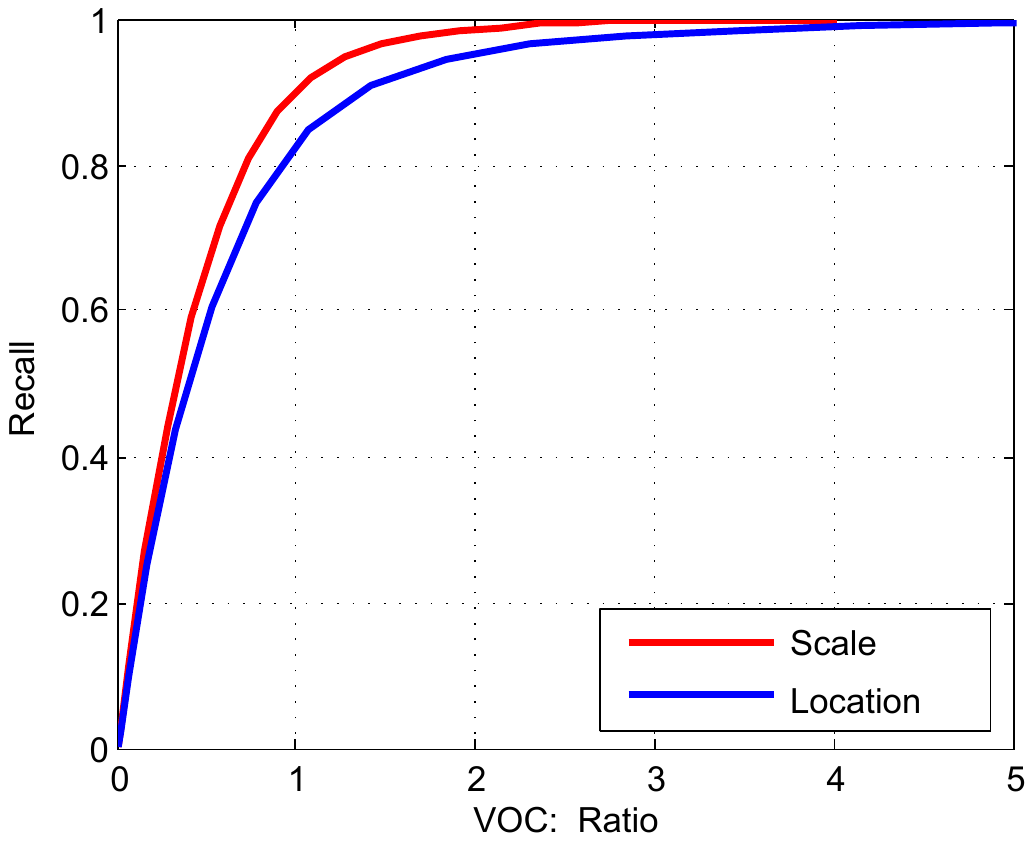}}
  \hspace{0.1in}
  \subfigure[Performance on COCO]{
    \label{fig:subfig:b} 
    \includegraphics[width=1.5in]{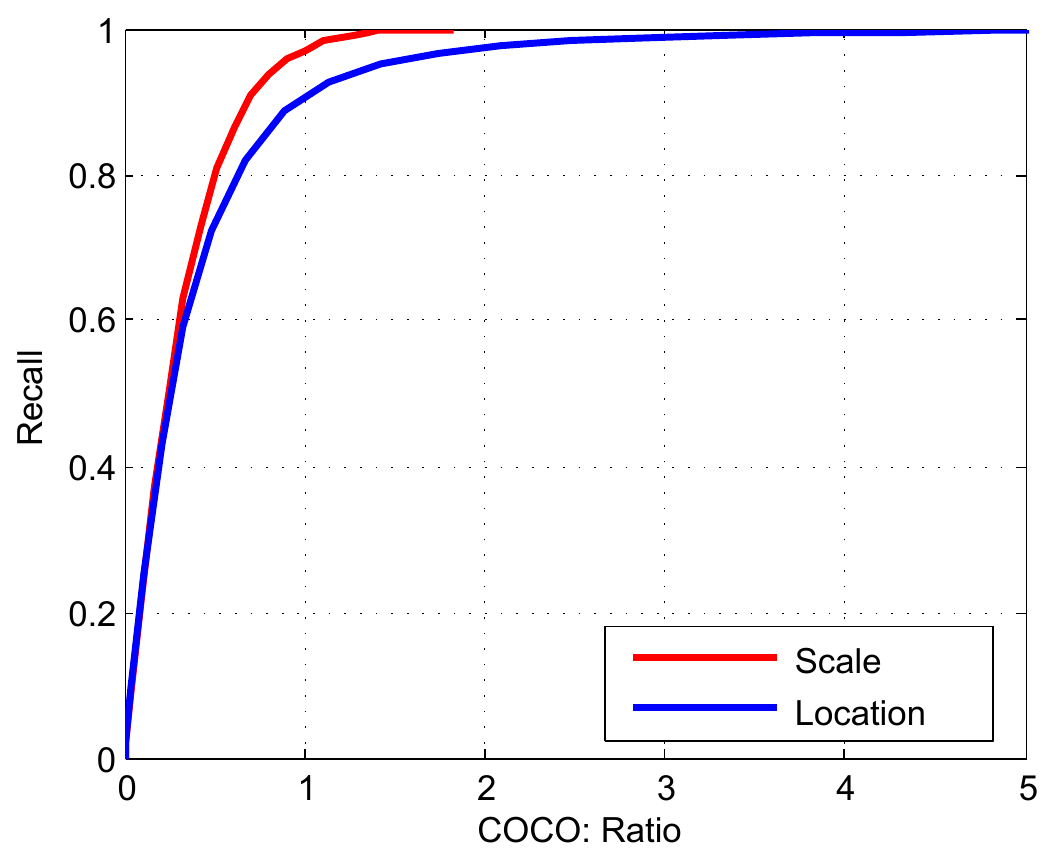}}
      \hspace{0.1in}
  \caption{Performance of $S^2AP$ on $VOC 2007$ and $COCO 2014$.}
  \label{fig:sap} 
\end{figure}

Face detection is the specific task of generic object detection. The excellent performance on `scale' and `spatial' of $S^2AP$ largely depends on the unified appearance of human face. In order to verify scalability of $S^2AP$, we perform experiments on popular generic object datasets Pascal VOC~\cite{everingham2010pascal} and COCO~\cite{lin2014microsoft}. Images from training sets of $VOC 2012+2007$ and $COCO 2014$ are used for training set and the testing is performed on testsets of $VOC 2007$ and $minival$ of $COCO 2014$. The configuration is same as above and the result of $S^2AP$ is shown in Figure~\ref{fig:sap}. Note that even though the aspect ratio is not uniform for generic objects, $S^2AP$ can also achieve high recall on both of scale and location. The performance in $COCO 2014$ is higher because of the concentrated distribution of object scale. Robustness of $S^2AP$ makes it possible to be embedded into FCN-based methods with no hesitate.

\section{Conclusion}
In this paper, we decompose the face searching space into two orthogonal directions, `scale' and `spatial'. A novel method named scale estimation and spatial attention proposal ($S^2AP$) is proposed to pay attention to some specific scales and valid locations in image pyramid.
Additional, we adopt a masked-convolution operation to accelerate FCN based on the attention result.
Experimental results show that our algorithm achieves new state-of-the-art while greatly accelerate FCN-based methods such as RPN.
$S^2AP$ and masked-convolution can dramatically speed up RPN by $4\times$ on average. Moreover, `scale' and `spatial' information estimated from the high-level representation by robust $S^2AP$ can benefit other tasks based on FCN.

\section{Acknowledgements}
This work is supported by the National Natural Science Foundation of China (No. 61472023) and Beijing Municipal Natural Science Foundation (No. 4182034).

{\small
\bibliographystyle{ieee}
\bibliography{egbib}
}

\end{document}